\documentclass[journal]{IEEEtran}

\usepackage{times}
\usepackage{epsfig}
\usepackage{graphicx}
\usepackage{amsmath}
\usepackage{amssymb}

\usepackage[mathscr]{eucal}
\usepackage{amsbsy}
\usepackage{bm}
\usepackage{fixltx2e}
\MakeRobust{\overrightarrow}
\include{macrofile}
\usepackage{booktabs}
\usepackage{mathtools}

\usepackage{amssymb}%,bbm}
\usepackage{amsmath,amsthm}
\usepackage{subfigure}
\usepackage{cite}
\usepackage{graphicx}
\usepackage{graphics}
\usepackage{color}
\usepackage{xspace}
\usepackage{bbm}
\usepackage{psfrag}
\usepackage{algorithmicx}
\usepackage{algorithm}
\usepackage{algpseudocode}
\usepackage{multirow}
\usepackage{array}
\usepackage{url}
\usepackage[normalem]{ulem}
\usepackage{graphicx}
\usepackage{floatflt,setspace}
\usepackage{algcompatible}
\algnewcommand\INPUT{\item[\textbf{Input:}]}%
\algnewcommand\OUTPUT{\item[\textbf{Output:}]}%

\usepackage{stackengine}
\def\delequal{\mathrel{\ensurestackMath{\stackon[1pt]{=}{\scriptstyle\Delta}}}}
\usepackage[breaklinks=true,bookmarks=false]{hyperref}

\newcommand{\argmin}{{\text{argmin}}}

\newtheorem{definition}{Definition}
\newtheorem{lemma}{Lemma}

\usepackage{array}
\usepackage{booktabs}

\begin{document}
%%%%%%%%% TITLE
\title{Principal Component Analysis with Tensor Train  Subspace}

\author{Wenqi Wang, Vaneet Aggarwal,  and Shuchin Aeron \thanks{W. Wang and V. Aggarwal are with Purdue University, West Lafayette, IN 47907, email: \{wang2041,vaneet\}@purdue.edu.  S. Aeron is with Tufts
		University, Medford, MA 02155, email: shuchin@ece.tufts.edu. 
		
		The work of W. Wang and V. Aggarwal was supported in part by the U.S. National Science Foundation
		under grant CCF-1527486. The work of S. Aeron was supported in part by NSF CAREER Grant \# 1553075 
}}

%% omit the following lines up until the closing ``}''.
%% Additional authors and addresses can be added with ``\and'',
%% just like the second author.
%% To save space, use either the email address or home page, not both
%\and

\maketitle
%\thispagestyle{empty}

%%%%%%%%% ABSTRACT
\begin{abstract}
Tensor train is a hierarchical tensor network structure that  helps alleviate the curse of dimensionality by parameterizing large-scale multidimensional data via a set of network of low-rank tensors. Associated with such a construction is a notion of Tensor Train subspace and in this paper we propose a TT-PCA algorithm for estimating this structured subspace from the given data. By maintaining low rank tensor structure, TT-PCA is more robust to noise comparing with PCA or Tucker-PCA. This is borne out numerically by testing the proposed approach on the Extended YaleFace Dataset B.  % It is theoretically proved that these TT-PCA methods achieve less storage requirements, and have computationally faster online implementation with improved classification performance. This is borne out by numerical simulations on the large MNIST handwritten digits data set. In summary, this paper provides a first evidence that tensor based methods can achieve better tradeoffs between computation, storage and error compared to traditional methods. 
\end{abstract}

%by exploiting efficient structures in multi-dimensional data. 
%can effectively capture the information in high order tensor data under large compression ratio. 
%This paper proposes a Tensor Train Principal Component Analysis (TT-PCA) algorithm to recover the subspace structure in the high order tensor data.  

%TT-PCA is robust to noise by maintaining the low rank structures in the high order tensor data. 

%A TT-PCA classifier that projects tensor data onto labeled tensor train subspace is proposed in this paper and the numerical results demonstrate that human faces under different illumination conditions fits  the tensor train subspace structure better as compared to the linear and multilinear subspace structures. 

%that  embeds a high order tensor into a low dimensional vector is proposed, which is used to give  a faster and storage efficient KNN classification algorithm in the embedded lower dimensional space. 

%Laplacian eigenmap embedding 

%The neighbor preserving embedding method helps preserve the local neighborhood structure of the manifold in the compressed vector data  and the numerical results on hand-written digits demonstrate that the embedded 2D images display better classification result as compared to the Tucker model based tensor embedding method.
%\vspace{-.25in}
\section{Introduction}
%\vspace{-.1in}
Robust feature extraction and dimensionality reduction are among the most fundamental problems in machine learning and computer vision. 
Assuming that the data is embedded in a low dimensional subspace, popular and effective methods for feature extraction and dimensionality reduction are the Principal Component Analysis (PCA) \cite{jolliffe2002principal, bishop2006pattern}, and the Laplacian eigenmaps \cite{belkin2003laplacian}. In this paper we consider \emph{structured subspace models}, namely tensor subspaces to further refine these subspace based approaches, with significant gains on processing and handling multidimensional data. 

%In many applications, there is nonlinearity in embedded manifold structure of the data that can be exploited by approaches like Locally Linear Embedding \cite{roweis2000nonlinear}, and Laplacian Eigenmap \cite{belkin2003laplacian}.  
%Most of the data around us are better represented as tensors to capture the correlations across different attributes. 
%Most of the real data around us is high dimensional data where tensors are better in representation to capture the correlations across different attributes. 
Most of the real-world data is multidimensional, i.e. it is a function of several independent variables, and typically represented by a multidimensional array of numbers. These arrays are often referred to as \emph{tensors}, terminology borrowed from multilinear algebra \cite{de2000multilinear}. For example, a color image can be considered as a third-order tensor, two of the dimensions (rows and columns) being spatial, and the third being spectral (color), while a color video sequence can be considered as an four-order tensor, time being the fourth dimension besides spatial and spectral. %This paper focuses on exploring novel approaches for linear and non-linear subspace embeddings, when the data is structured as a low tt-rank tensor\cite{oseledets2011tensor}. 

%The problem of finding low dimensional structure in high dimensional data has been widely applied in image processing, pattern recognition and machine learning. 
%PCA \cite{jolliffe2002principal, bishop2006pattern} is the most widely statistical method for dimensional reduction, data denoising and data analysis. 
%However,  in reality most of the data are multi-dimensional data, such as colorful images, videos, fMRI imaging and etc., where analyzing vectorized high dimensional data by PCA suffers from losing structure information of the high dimensional data, and this problem becomes more serious when applying PCA to do dimension reduction in high dimensional features. 

A very popular tensor representation format namely the Tucker format has shown to be useful for a variety of applications  \cite{de2000multilinear, donoho1995noising, lu2006multilinear, vasilescu2003multilinear, zeng2014multilinear}. However, for large tensors, Tucker representation can still be exponential in storage requirements. In  \cite{holtz2012manifolds} it was shown that hierarchical Tucker representation, and in particular Tensor Train (TT) representation can alleviate this problem. Nevertheless, the statistical and computational tradeoffs of such reduced complexity formats, such as the TT, have not been studied so far. 

%One of the key used tensor structure is the tucker tensor structure, where the data unfolded to each matrix unfold is low rank.

%The extension of  singular value decomposition to Tucker structure has been  investigated to learn the multilinear tensor subspace \cite{de2000multilinear,lu2011survey} and has been shown to be effective in dimensional reduction with low reconstruction error \cite{vasilescu2003multilinear, lu2006multilinear, zeng2014multilinear}. 
%However,  the Tucker tensor structure is subject to curse of dimensionality, which can be alleviated by the Tensor Train representation of the tensor structure \cite{holtz2012manifolds}. The data with Tensor Train (TT) decomposition can be represented as a product of third order tensors, denoted as matrix product state (MPS).

In this paper, we begin by showing that TT decompositions are associated with a structured subspace model, namely the Tensor Train subspace. Based on this model, the problem of finding the Tensor Train subspace and the representation of the data is formulated as an extension of the Tucker and traditional PCA based technique. An algorithm to solve this non-convex problem is provided, and is referred to as TT-PCA.  We show that if the data admits a TT representation, then TT-PCA significantly reduces storage and complexity as compared to the standard PCA and the Tucker-PCA (T-PCA). We use TT-PCA for classification on Extended YaleFace Dataset B \cite{georghiades2001few,lee2005acquiring}, where different images of 38 humans are classified. We see that TT-PCA is able to exploit the structure better than the standard PCA and the T-PCA approaches and achieves the lowest classification error at a lower amount of compressed data dimension.

The rest of the paper is organized as follows. The technical notations and definitions are introduced in Section \ref{NP}.  The  tensor train subspace (TT-subspace) model is described in Section \ref{TTS}, and TT-PCA algorithm is proposed in Section \ref{sec:TTPCA}. %In Section \ref{NPE}, the tensor train neighbor preserving embedding (TT-NPE) problem is formulated, and an efficient algorithm is proposed. 
Section \ref{simu} provides the numerical results for the proposed algorithms on Yale Face and MNIST databases. Finally, Section \ref{concl} concludes the paper. 
%, we conduct numerical experiment on face denoising, face compression, face classification to verify our results in tensor train PCA. A simulation on digit recognition is then conducted verify the proposed embedding algorithm. In the last section, we give the concluding remarks and potential future directions.

%\vspace{-.1in}
\section{Notations and Preliminaries}\label{NP}
%\vspace{-.1in}
In this section, we introduce the notations that will be extensively used in this paper. Vectors and matrices are represented by bold face lower letters (e.g. ${\bf x}$) and bold face capital letters (e.g. ${\bf X}$), respectively. An identity matrix of $r$ rows is denoted as ${\bf I}^r$. Transpose of matrix ${\bf X}$ is denoted by ${\bf X}^T$.  
An $n^{\text{th}}$ order tensor is denoted by calligraphic letter ${\mathscr{X}} \in \mathbb{R}^{I_1 \times I_2 \times ... \times I_n}$, where $I_{i: i=1,2,..., n}$ is the dimension along the $i^\text{th}$ mode. An entry inside a tensor $\mathscr{X}$ is represented as $\mathscr{X}(i_1, i_2,\cdots, i_n)$, where $i_{k: k=1,2,.., n}$ is the location index along the $k^{\text{th}}$ mode. 
A colon is applied to represent all the elements of a mode in a tensor,  e.g. $\mathscr{X}(:, i_2,\cdots, i_n)$ represents the fiber along mode $1$ and $\mathscr{X}[:, :, i_3, i_4,\cdots, i_n]$ represents the slice along mode $1$ and mode $2$ and so forth. ${\bf V}(\cdot)$ is a tensor vectorization operator such that  $\mathscr{X} \in \mathbb{R}^{I_1 \times \cdots \times I_n }$ maps to a vector ${\bf V}({\mathscr{X}}) \in \mathbb{R}^{I_1  \cdots  I_n }$.

% and give definition that inherits from \cite{wang2016tensor}.
%Vectors and matrices are represented by bold face lower letters e.g.${\bf x,y,z,...}$ and bold face capital letters, e.g.${\bf X, Y, Z,...}$, respectively. 
%An identity matrix is denoted as ${\bf I}$ through out the paper.
%Tensors with order more than three are represented by bold face calligraphic letters $\bf \mathscr{X}, \mathscr{Y}, \mathscr{Z}$. 
%Typically, a $n^\text{th}$ order tensor is represented by ${\bf \mathscr{X}} \in \mathbb{R}^{I_1 \times I_2 \times ... \times I_n}$, where $I_{i: i=1,2,..., n}$ is the dimension along the $i^\text{th}$ mode. 
%Entries inside a tensor are denoted by Matlab notation where a colon is applied to represent all elements of a mode. 
%For instance, $\mathscr{X}(i_1, i_2,..., i_n)$ represents an entry in tensor $\mathscr{X}$ at location $i_{k: k=1,2,.., n}$ along the $k^{\text{th}}$ mode,  
% $\mathscr{X}[:, i_2,..., i_n]$ represents the fiber along mode-$1$, $\mathscr{X}[:, :, i_3, i_4,..., i_n]$ represents the slice along mode-$1$ and mode-$2$ and so forth.

% ${\bf V}(\cdot)$ is a tensor vectorization operator such that for a set of tensors $\mathscr{X} \in \mathbb{R}^{I_1 \times \cdots \times I_n \times N}$, ${\bf V}(\mathscr{X}) ={\bf M} \in \mathbb{R}^{(I_1 \cdots I_n) \times N}$. ${\bf T}(\cdot)$ is a vector tensorization operator, which is the reverse operator of ${\bf V}(\cdot)$,  such that ${\bf T}({\bf M}) = \mathscr{X} \in \mathbb{R}^{I_1 \times \cdots \times I_n \times N}$.

%Let $\mathscr{X}\in\mathbb{R}^{I_1 \times \cdots I_n}$ be a $n^\text{th}$ order tensor. 
Under tucker format \cite{cichocki2016low}, any entry insider a tensor is represented by the Tucker Decomposition 

\vspace{-.2in}
\begin{equation} \label{eq: Tucker}
\begin{split}
&\mathscr{X}(i_1,\cdots, i_n)  \nonumber \\
= &\sum_{j_1=1}^{r_1} \cdots \sum_{j_n=1}^{r_n} \mathscr{C}(j_1,\cdots, j_n) {\bf U}_1(i_1, j_1) \cdots {\bf U}_n(i_n, j_n),
\end{split}
\end{equation}
where $\mathscr{C} \in \mathbb{R}^{r_1 \times \cdots \times r_n}$ is the core tensor and ${\bf U}_i\in \mathbb{R}^{I_i\times r_i}$ are the set of orthonormal linear transformation that  defines the tucker structure. The Tucker-Rank is denoted by the vector of ranks $(r_1, \cdots, r_n)$ in the Tucker Decomposition. %Under tensor notation, \eqref{eq: Tucker} is equivalent to 
%\begin{equation}
%\mathscr{X} = \mathscr{C} \times_1 {\bf U}_1 \cdots \times_n {\bf U}_n,
%\end{equation}
The multilinear subspace is defined by the span of a given set of core tensors after the set of linear transformation given by ${\bf U}_i$. In this paper, we refer the method to recover the multilinear subspace, or the Tucker Subspace, as Tucker PCA (T-PCA).

Tensor train decomposition \cite{oseledets2011tensor, holtz2012manifolds} is a tensor factorization method that any elements inside a tensor $\mathscr{X} \in \mathbb{R}^{I_1 \times\cdots\times I_n}$, denoted as $\mathscr{X}(i_1, i_2,\cdots, i_n)$, is represented by
\vspace{-.1in}
\begin{equation}
\begin{split}
&\mathscr{X}(i_1,\cdots, i_n) \nonumber \\ 
&= {\bf U}_1(i_1,:) \mathscr{U}_2(:, i_2, :) \cdots \mathscr{U}_{n-1}(:, i_{n-1}, :){\bf U}_n(:, i_n),
\end{split}
\end{equation}
where ${\bf U}_1\in\mathbb{R}^{I_1 \times r_1}$, ${\bf U}_n\in\mathbb{R}^{r_{n-1} \times I_n}$ are the boundary matrices and $\mathscr{U}_i \in \mathbb{R}^{r_{i-1} \times I_i \times r_i}, i=2,\cdots, n-1$ are middle decomposed tensors. 
Without loss of generality, we can define ${\mathscr{U} }_1\in\mathbb{R}^{1\times I_1 \times r_1}$ as the tensor representing ${\bf U}_1$, and $\mathscr{ U}_n\in\mathbb{R}^{r_{n-1} \times I_n\times 1}$ representing ${\bf U}_n$. The TT-Rank of a tensor is denoted by the vector of ranks $(r_1, \cdots, r_{n-1})$ in the tensor train decomposition. We let $r_0=1$.  
We note that the representation of a tensor as the product of tensors is called the Matrix Product State Structure, and the different ${\mathscr{U} }_i$ are called Matrix Product States (MPS)\cite{perez2006matrix}.

%Under tensor train formation \cite{oseledets2011tensor, holtz2012manifolds}, any entry inside the tensor, denoted as $\mathscr{X}( i_1,\cdots, i_n )$, is represented by
%\begin{equation}
%\mathscr{X}( i_1,\cdots, i_n ) = {\bf U}_1(i_1) {\bf U}_2(i_2) \cdots {\bf U}_n(i_n),
%\end{equation}
%where ${\bf U}_1(i_1) \in \mathbb{R}^{1 \times r_1}$, ${\bf U}_n(i_n) \in \mathbb{R}^{r_n \times 1}$ are vectors and ${\bf U}_{i:i=2,\cdots n-1} (i_i) \in \mathscr{R}^{r_{i-1} \times r_i}$ are matrices. This is equivalent to tensor notation \cite{wang2016tensor}
%\begin{equation}\label{eq: TT}
%\begin{split}
%&\mathscr{X}( i_1,\cdots, i_n ) \\
%= & \mathscr{U}_1(1,i_1, :)\mathscr{U}_2(: ,i_2, :) \cdots \mathscr{U}_{n-1}(: ,i_{n-1}, :)  \mathscr{U}_n(:,i_n, 1),
%\end{split}
%\end{equation}
%where $\mathscr{U}_1 \in\mathbb{R}^{1 \times I_1 \times r_1}$, $\mathscr{U}_n \in\mathbb{R}^{r_{n-1} \times I_n \times 1}$ and $\mathscr{U}_i\in\mathbb{R}^{r_{i-1} \times I_i \times r_i}$ are $3$-mode tensor, or MPS. 

We next define the mode-$i$ unfolding of a tensor as follows. 
\vspace{-.05in}
{\definition (Mode-$i$ unfolding \cite{cichocki2014era}) Let $\mathscr{X} \in \mathbb{R}^{I_1 \times \cdots \times I_n}$ be a $n$-mode tensor. Mode-$i$ unfolding of $\mathscr{X}$, denoted as $\mathscr{X}_{[i]}$, matrized the tensor $\mathscr{X}$ by putting the $i^{\text{th}}$ mode in the matrix rows and remaining modes with the original order in the columns such that
	\begin{equation}
	\mathscr{X}_{[i]} \in \mathbb{R}^{I_i \times (I_1\cdots I_{i-1}I_{i+1}\cdots I_n)} .
	\end{equation}
}
\vspace{-.25in}
%{\definition (Mode-$i$ matrization \cite{cichocki2014era}) Let Let $\mathscr{X} \in \mathbb{R}^{I_1 \times \cdots \times I_n}$ be a $n$-mode tensor. Mode-$i$ unfolding of $\mathscr{X}$, denoted as $\mathscr{X}_{<i>}$, matrized the tensor $\mathscr{X}$ by putting the first $1,2,\cdots, i$ mode in the matrix rows and remaining $i+1,\cdots, n$ order in the columns such that
%	\begin{equation}
%	\mathscr{X}_{<i>} \in \mathbb{R}^{(I_1 \cdots I_i) \times (I_{i+1}\cdots I_n)} .
%	\end{equation}} 

Next, we define the notion of left-unfolding for third order tensors. 
\vspace{-.1in}
\begin{definition}(Left Unfolding \cite{holtz2012manifolds}) Let $\mathscr{X} \in \mathbb{R}^{r_{i-1} \times I_i \times r_i}$ be a third order tensor, the left unfolding is the matrix obtained by taking the first two modes indices as rows indices and the third mode indices as column indices such that
%	\begin{equation}
$	{\bf L}(\mathscr{X}) \in \mathbb{R}^{(r_{i-1}I_i ) \times r_i}$, and is given as ${\bf L}(\mathscr{X})  = 	(\mathscr{X}_{[3]})^T$. 
%	\end{equation}
%	and the mapping relation is
%	\begin{equation}
%	\begin{split}
%	&{\bf L}(\mathscr{X})[k_0 + (i_1-1)r_0, k_1] = \mathscr{X}[k_0, I_1, k_1], \\
%	&\forall_{k_0\in[1, r_0], i_1\in[1, I_1], k_1\in[1, r_1]}.
%	\end{split}
%	\end{equation}
	
	(Left Refolding) Left refolding operator ${\bf L}^{-1}$ is the reverse operator of left unfolding ${\bf L}$,  which reshapes a $\mathbb{R}^{(r_{i-1}I_i) \times r_i}$ matrix to a $\mathbb{R}^{r_{i-1} \times I_i \times r_i}$ tensor. % such that
%	\begin{equation}
%	{\bf L}^{-1}(\mathbb{R}^{(r_{i-1}I_i) \times r_i}) \rightarrow \mathbb{R}^{r_{i-1} \times I_i \times r_i}.
%	\end{equation}

Similar to left unfolding and refolding, right unfolding is ${\bf R}(\mathscr{X}) = \mathscr{X}_{[1]} \in \mathbb{R}^{r_{i-1} \times (I_ir_i)}$.
% and right unfolding ${\bf R}^{-1}$ is the reverse operator of right unfolding that reshapes a $\mathbb{R}^{r_{i-1} \times (I_1r_i)}$ tensor to a $\mathbb{R}^{r_{i-1} \times I_i \times r_i}$.
\end{definition}

\vspace{-.15in}
{\definition(Tensor Connect Product \cite{wang2016tensor})
Let $\mathscr{U}_i \in \mathbb{R}^{r_{i-1} \times I_i \times r_i}, i=1,\cdots, n$ be $n$ third order tensors. The tensor connect product $\mathscr{U}_1 \cdots \mathscr{U}_n $ is defined as
\vspace{-.05in}
\begin{equation}
\mathscr{U} =\mathscr{U}_1 \cdots \mathscr{U}_n \in \mathbb{R}^{r_0 \times (I_1\cdots I_n) \times r_n},
\vspace{-.1in}
\end{equation}
where for any two adjacent third-order tensor, the tensor connect product satisfies
\vspace{-.1in}
\begin{equation}
\begin{split}
\mathscr{U}_j\mathscr{U}_{j+1} &\in \mathbb{R}^{r_{j-1} \times (I_jI_{j+1}) \times r_{j+1}} \\
&= {\bf L}^{-1} \left( {\bf I}^{(I_{j+1})} \otimes {\bf L}(\mathscr{U}_j) \times {\bf L}(\mathscr{U}_{j+1}) \right).
\end{split}
\vspace{-.1in}
\end{equation}
Tensor connect product is the tensor product for third order tensors, and matrix product for second order tensors (matrices). 
}
\vspace{-.1in}

%With the tensor connect product, \eqref{eq: TT} gives a more concise tensor for
%\begin{equation}
%{\bf V}(\mathscr{X}) = {\bf V}(\mathscr{U}_1\cdots \mathscr{U}_n).
%\end{equation}

%Other commonly used $3$-mode tensor operation and concept, including left unfolding (refolding), mode-$n$ unfolding are defined as follows. 

% ==========   Definition 1    =======================

% ===========  Definition 2   ===========================

% ==========  Definition 3  ==========================
%{\definition (Tensor Train Decomposition \cite{oseledets2011tensor, holtz2012manifolds}) Let $\mathscr{U} \in \mathbb{R}^{I_1 \times \cdots \times I_n}$ be an arbitrary tensor. For any entry inside the tensor, denoted as $\mathscr{U}( x_1,\cdots, x_n )$, a TT decomposition gives
%\begin{equation}
%\mathscr{U}( x_1,\cdots, x_n ) = {\bf U}_1(x_1) {\bf U}_2(x_2) \cdots {\bf U}_n(x_n),
%\end{equation}
%where ${\bf U}_1(x_1) \in \mathbb{R}^{1 \times r_1}$, ${\bf U}_n(x_n) \in \mathbb{R}^{r_n \times 1}$ are vectors and ${\bf U}_{i:i=2,\cdots n-1} (x_i) \in \mathscr{R}^{r_{i-1} \times r_i}$ are matrices. }

%Based on the tensor train definition, a tensor connect product is further introduced such that tensor train format can be generalized as the product of a sequence of $3$-mode tensor.
% ==========   Definition 4    ===========================

\section{Tensor Train Subspace (TTS)}\label{TTS}
%\vspace{-.1in}
A tensor train subspace, ${\bf S_{\text{TT}}} \subseteq \mathbb{R}^{I_1  \cdots  I_n}$,  is defined as the span of a matrix that is generated by the left unfolding of a tensor, such that 
\begin{equation}
\begin{split}
{\bf S_{\text{TT}}} &\delequal \text{span}({\bf L}(\mathscr{U}_1\mathscr{U}_2...\mathscr{U}_n)) \\
&= \{ {\bf L}(\mathscr{U}_1\mathscr{U}_2...\mathscr{U}_n){\bf a} | \forall {{\bf a} \in \mathbb{R}^{r_n }} \}.
\end{split}
\end{equation}

We note that a tensor subspace is determined by  $\mathscr{U}_1, \mathscr{U}_2, \cdots, \mathscr{U}_n$, where $\mathscr{U}_i \in \mathbb{R}^{r_{i-1} \times I_i \times r_i}$, $r_0 =1$ \cite{wang2017tensor}. In a special case when $n=1$, the proposed tensor train subspace is reduced to the linear subspace model under matrix case.

The next result shows that for the given $\mathscr{U}_1, \mathscr{U}_2, \cdots, \mathscr{U}_n$, ${\bf S_{\text{TT}}} $ is a subspace.
%=================   Lemma    ======================================
%\vspace{-.1in}
{\lemma \label{LemmaSP} (Subspace Property \cite{wang2017tensor}) ${\bf S_{\text{TT}}} $ is a subspace of $\mathbb{R}^{I_1  \cdots  I_n}$ for given $\mathscr{U}_1, \mathscr{U}_2, \cdots, \mathscr{U}_n$.}

We next give some properties of the TT decomposition that will be used in this paper. 

{\lemma (Left-Orthogonality Property \cite[Theorem 3.1]{holtz2012manifolds})
For any tensor 	$\mathscr{X}$ of TT-rank ${\mathbf r}$, the TT decomposition  can be chosen such that  ${\bf L}(\mathscr{U}_i)$ are left-orthogonal for all $i=1, \cdots n$, or ${\bf L}(\mathscr{U}_i)^T {\bf L}(\mathscr{U}_i) = {\bf I}_{r_i} \in \mathbb{R}^{r_i \times r_i}$.
	}
	
We next show that if ${\bf L}(\mathscr{U}_i)$ is left-orthogonal for all $i=1, \cdots, n$, then  ${\bf L}(\mathscr{U}_1\cdots \mathscr{U}_n)$ is left-orthogonal.

{\lemma (Left-Orthogonality of Tensor Connect Product \cite{wang2017tensor}) 
	 If ${\bf L}(\mathscr{U}_i)$ is left-orthogonal for all $i=1, \cdots, n$, then  ${\bf L}(\mathscr{U}_1\cdots \mathscr{U}_j)$ is left-orthogonal for all $1\le j\le n$. \label{tcplo}
}

Thus, we can without loss of generality, assume that ${\bf L}(\mathscr{U}_i)$ are left-orthogonal for all $i$. Then, the projection of a data point ${\bf y}\in \mathbb{R}^{r_n }$ on the subspace ${\bf S_{\text{TT}}} = \text{span}({\bf L}(\mathscr{U}_1\mathscr{U}_2...\mathscr{U}_n))$  is given by ${\bf L}(\mathscr{U}_1\cdots \mathscr{U}_n)^T {\bf y}$.
%\vspace{-.05in}
\section{Tensor Train PCA}\label{sec:TTPCA}
%\vspace{-.05in}
Given a set of tensor data $\mathscr{X}_{i}\in \mathbb{R}^{I_1 \times \cdots \times I_n}$,  $i=1,\cdots, N$,  we intend to find $r_n$ principal vectors  that convert a set of observations of possibly correlated variables into a set of values of linearly uncorrelated variables. The $r_n$ principal vectors can be stacked as a matrix
${\bf L}(\mathscr{U}_1\mathscr{U}_2\cdots\mathscr{U}_n)\in \mathbb{R}^{I_1  \cdots  I_n\times r_n}$ such that $\mathscr{U}_i\in  \mathbb{R}^{r_{i-1}\times I_i \times \times r_i}$, with $r_0=1$. The objective of Tensor Train PCA (TT-PCA) is to find such $\mathscr{U}_1, \mathscr{U}_2, \cdots, \mathscr{U}_n$ such that the distance of the points from the TTS formed by $\mathscr{U}_1, \mathscr{U}_2, \cdots, \mathscr{U}_n$ is minimized. We note that for $n=1$, this is the same objective as that for standard PCA \cite{vidal2005generalized}.

Given  $N$ data points  $\mathscr{X}_{i}\in \mathbb{R}^{I_1 \times \cdots \times I_n}$,  $i=1,\cdots, N$, let ${\bf D}\in \mathbb{R}^{I_1  \cdots  I_n\times N}$ be the matrix that concatenates the $N$ vectorizations such that the $i^{\text{th}}$ column of ${\bf D}$ is  ${\bf V}({\mathscr{X}}_i)$. The goal then is to find $\mathscr{U}_1, \mathscr{U}_2, \cdots, \mathscr{U}_n$ such that the distance of points from the subspace is minimized. More formally, we wish to solve the following problem,
\begin{equation} \label{eq: TTPCA_Formulation}
\min_{\mathscr{U}_{i}, i=1,\cdots, n, {\bf A}}  \| {\bf L}(\mathscr{U}_1 \cdots \mathscr{U}_n){\bf A}- {\bf D}\|_F^2.
\end{equation}

\subsection{Algorithm}
\vspace{-.05in}
This optimization problem in \eqref{eq: TTPCA_Formulation} is a non-convex problem. We however note that the problem is convex w.r.t. each of the variables ($\mathscr{U}_{i}, i=1,\cdots, n, {\bf A}$) when the rest are fixed. Thus, one approach to solve the problem is to alternatively minimize over the variables when the rest are fixed.

%Under tensor train subspace model, a tensor data is a tensor $\mathbb{R}^{I_1 \times \cdots \times I_n}$ and we are given a set of $N$ tensor data, denoted as $\mathscr{X} \in \mathbb{R}^{I_1 \times \cdots \times I_n \times N}$.  The goal is to find a set of bases $\mathscr{U}_{i:i=1,..., n}$ and ${\bf A} \in \mathbb{R}^{r_n \times N}$ such that
%\begin{equation} \label{eq: TTPCA_Formulation}
%\min_{\mathscr{U}_{i:i=1,.., n}, {\bf A}}  \| {\bf L}(\mathscr{U}_1 \cdots %\mathscr{U}_n){\bf A}- {\bf V}(\mathscr{X})\|_F^2.
%\end{equation}

%To avoid the multiple solutions given by \eqref{eq: TTPCA_Formulation} , a set of unitary constraints are added in \eqref{eq: TTPCA_Formulation} such that \eqref{eq: TTPCA_Formulation} becomes
%\begin{equation}
%\begin{split}
%&\min_{\mathscr{U}_{i:i=1, \cdots, n}, {\bf A}}  \| {\bf L}(\mathscr{U}_1 \cdots \mathscr{U}_n){\bf A}- {\bf V}(\mathscr{X})\|_F^2 \\
%s.t. & \forall_{i:i=1,\cdots, n} {\bf L}(\mathscr{U}_i)^\top {\bf L}(\mathscr{U}_i) = {\bf I} \in \mathbb{R}^{r_i \times r_i}
%\end{split}
%\end{equation}

In this paper, we propose an alternate approach that is based on successive SVD-algorithm  for computing TT Decomposition in \cite{holtz2012manifolds}. The algorithm steps are given in Algorithm \ref{TTPCA_Algo}. The algorithm steps assume that rank vector is not known, and estimates the ranks based on thresholding singular values. However, if the ranks are known, the threshold will be at the $r_i$ number of singular values rather than at $\tau$ fraction of the maximum singular value. The proposed algorithm goes from left to right and find the different $\mathscr{U}_{i}$s. We note that this algorithm extends computing TT Decomposition in \cite{holtz2012manifolds} by thresholding over the singular values, which tries to find the low rank approximation since the data is not exactly low rank. Such approaches for thresholding singular values for data approximation to low rank have been widely used for matrices \cite{cai2010singular,donoho1995noising}.

The advantage of the approach include the following: (i) There are no iterations like in Alternating Minimization based approach, and the complexity is low. (ii) The obtained ${\bf L}(\mathscr{U}_{i})$ is left-orthogonal for all $i = 1, \cdots, N$. Due to this property, we have by Lemma \ref{tcplo} that ${\bf L}(\mathscr{U}_1\cdots \mathscr{U}_n)$ is left-orthogonal. Thus, the projection of a data point ${\mathscr D} \in  \mathbb{R}^{I_1 \times \cdots \times I_n}$ onto the TT subspace formed is $({\bf L}(\mathscr{U}_1\cdots \mathscr{U}_n))^T {\bf V}(\mathscr D)$ .

%We propose a Tensor Train Principle Component Analysis (TT-PCA) algorithm by taking orders to apply Singular Value Decomposition (SVD) and threshold singular values to recover the MPS, and thus tensor train subspace. The TT-PCA algorithm is described in Algorithm \ref{TTPCA_Algo}.
\begin{algorithm}
   \caption{Tensor Train Principle Component Analysis (TT-PCA) Algorithm}
   \begin{algorithmic}[1]
   \INPUT   $N$ tensors $\mathscr{X}_i \in \mathbb{R}^{I_1\times I_2\times \cdots \times I_n }$, $i=1, \cdots, N$, threshold parameter $\tau$ %, rank vector ${\bf r} = (r_1, \cdots r_n)$
    \OUTPUT  Decomposition for tensor train subspace $\mathscr{U}_1, \mathscr{U}_2, \cdots, \mathscr{U}_n$ and the representation ${\bf A}$
    \STATE Form $\mathscr{Y}$ as an order $n+1$ tensor s.t. $\mathscr{Y} \in \mathbb{R}^{I_1\times I_2\times \cdots \times I_n \times N}$, which is formed by concatenating all data points $\mathscr{X}_i$ in the last mode. 
    \STATE Set ${\bf X}_1$ to be the $\mathscr{Y}_{[1]} \in \mathbb{R}^{I_1 \times (I_2 \cdots I_nN)}$ and apply SVD to ${\bf Y}_1$ such that ${\bf Y}_1 ={\bf U}_1{\bf S}_1{\bf V}_1^\top$. Threshold singular values in ${\bf S}_1$ by maintaining the singular value larger than $\tau \sigma_{\max_1}$, where $\sigma_{\max_1}$ is the largest singular value of ${\bf S}_1$,   to get $\tilde{{\bf S}}_{1}$ and the number of non-zero singular values in $\tilde{{\bf S}}_{1}$ as $r_1$, calculate ${\bf X}_2 = {\tilde {\bf S}}_{1} {\bf V}^\top$ and set $\mathscr{U}_1 = {\bf L}^{-1}({\bf U}_1) \in \mathbb{R}^{1 \times I_1 \times r_1}$.
    \FOR{ $i = 2$ to $n$}
    	\STATE Reshape ${\bf X}_i \in \mathbb{R}^{r_{i-1} \times (I_{i}\cdots I_nN)}$ to ${\bf Y}_i \in \mathbb{R}^{(r_{i-1}I_{i}) \times (I_{i} \cdots I_nN)}$ and apply SVD to ${\bf Y}_i$ such that ${\bf Y}_i = {\bf U}_i {\bf S}_i {\bf V}_i^\top$
	\STATE Threshold singular values in ${\bf S}_i$ by maintaining the singular value larger than $\tau \sigma_{\max_i}$ to get ${\tilde{\bf S}}_{i}$ and the number of non-zero singular values in ${\tilde{\bf S}}_{i}$ as $r_i$.
	\STATE Set $\mathscr{U}_i = {\bf L}^{-1}({\bf U}_i) \in \mathbb{R}^{r_{i-1} \times I_i \times r_i}$ and ${\bf X}_{i+1} ={\tilde {\bf S}_{i}}{\bf V}_i^\top$
    \ENDFOR
    \STATE Set ${\bf A} = {\bf X}_{n+1}$
\end{algorithmic}
\label{TTPCA_Algo}
\end{algorithm} 

%{\lemma  \label{LemmaOP} (Orthonormal Property) The left-unfolding of tensor connect product that generated by first $i$ recovered MPS , denoted as ${\bf L}(\mathscr{U}_1\cdots \mathscr{U}_{i:\forall_{i=1,\cdots, n}})$, is a unitary matrix. Let ${\bf B}_i = {\bf L}(\mathscr{U}_1 \cdots \mathscr{U}_{i}), \forall_{i=1,\cdots, n}$, then
%\begin{equation} \label{eq: OP}
%\forall_{i=1,\cdots, n}, \quad {\bf B}_i^\top {\bf B}_i = {\bf I} \in \mathbb{R}^{r_i \times r_i}
%\end{equation}
%}
%\proof
%Proof is in Appendix \ref{ProofOP}.
%\endproof

\vspace{-.05in}
\subsection{Classification Using TT-PCA}
\vspace{-.05in}
In order to use TT-PCA for classification, we assume that we have $N_\text{tr}$ data points $\mathscr{X}_i \in \mathbb{R}^{I_1\times I_2\times \cdots \times I_n }$, $i=1, \cdots, N_\text{tr}$ for training, each having label $l_i\in \{1, \cdots, C\}$ that identify the association of the data points to  the $C$ classes, and let $N_\text{te}$ data points as test data points that we wish to classify into the $C$ classes. The first step is to perform TT-PCA for each of the $C$ classes based on the data points that have that particular label among the $N$ training data points. Let the corresponding $\mathscr{U}_{i:i=1,\cdots, n}$ for class $j$ be denoted as $\mathscr{U}_{i:i=1,\cdots, n}^{(j)}$. Further, let ${\bf U}^{(j)} = {\bf L}(\mathscr{U}^{(j)}_1 \cdots \mathscr{U}^{(j)}_n)$. For a  data point in the testing set  $\mathscr{Y} \in \mathbb{R}^{I_1 \times \cdots I_n}$, we wish to decide its label based on its distance to the subspace. Thus, the assigned label is given by 
%
%TT-PCA can be applied for classification where a vectorized tensor data is projected onto a set of tensor train subspace and labeled by the subspace label where the projection residual is minimized. Let $\mathscr{U}_{i:i=1,\cdots,n}^{(j)}$ be the set of MPS of the tensor train subspace with the label $j$, ${\bf U}^{(j)} = {\bf L}(\mathscr{U}^{(j)}_1 \cdots \mathscr{U}^{(j)}_n)$ and ${\bf y} = {\bf V}(\mathscr{Y} \in \mathscr{R}^{I_1 \times \cdots I_n})$ be the vectorized tensor data, then the predicted label for tensor data $\mathscr{Y}$ is made by solving
\vspace{-.1in}
\begin{equation}\label{eq: pred}
\text{Label}(\mathscr{Y}) = \argmin_{j= 1, \cdots, C} \|{\bf U}^{(j)}{{\bf U}^{(j)}}^\top {\bf V}(\mathscr{Y}) -{\bf V}(\mathscr{Y})\|_2^2.
\vspace{-.1in}
\end{equation}

\vspace{-.05in}
\subsection{Storage and Computation Complexity}
\vspace{-.05in}
In this subsection, we will give the amount of storage needed to store the subspace, and complexity for doing TT-PCA and classification based on TT-PCA. For comparisons, we consider the standard PCA and Tucker based PCA (T-PCA) algorithm \cite{de2000multilinear}. We  let $d=I_1\cdots I_n$ be the dimension of each vectorized $n^{\text{th}}$ order tensor data. Suppose we have $N$ data points. We assume that $I_1 = \cdots = I_n$. Further, rank for PCA is chosen to be $r$, rank in each unfold for T-PCA is assumed to be $r$, and the ranks $r_i=r$ for $i\ge 1$ are chosen for TT-PCA. We note that ranks in each decomposition have a different interpretation and not directly comparable.

\noindent {\bf Storage of subspace:} Under PCA model, the storage needed is for a $d\times r$ matrix which is left-orthogonal, and thus
\vspace{-.1in}
\begin{equation}
\text{dim(PCA)} = dr -r(1+r)/2,
\vspace{-.05in}
\end{equation}
where the $r(1+r)/2$ component is saved in storage as a result of orthonormal property of the PCA bases. %The corresponding compression ratio is
%\begin{equation}
%\rho_{\text{PCA}} = \frac{\text{dim(PCA)}}{Nd}
%\end{equation}

Under T-PCA model,  $n$ linear transformations and $r$ core tensors need to be stored, and thus
\vspace{-.1in}
\begin{equation}
\text{dim(T-PCA)} = r^{n+1} + n \left( d^{\frac{1}{n}}r - r(1+r)/2  \right),
\vspace{-.05in}
\end{equation}
where $r^{n+1}$ is the storage for $r$ cores, each  $\in \mathbb{R}^{r\times \cdots \times r}$, and $n(d^{\frac{1}{n}}r - r(1+r)/2 )$ is the storage for $n$ linear transformations.  $n r(1+r)/2 $ amount of storage is saved due to the orthonormal property of the linear transformation matrices. %Thus, the compression ratio is
%\begin{equation}
%\rho_{\text{T-PCA}} = \frac{\text{dim(T-PCA)}}{Nd}
%\end{equation}

Under TT-PCA model, we need to store $\mathscr{U}_1, \cdots,  \mathscr{U}_n$ which are all left-orthogonal, and thus
\vspace{-.1in}
\begin{equation}
\text{dim(TT-PCA)} = d^{\frac{1}{n}} r(r(n-1)+1) -r(1+r)n/2,
\vspace{-.05in}
\end{equation}
where $\mathscr{U}_1$ takes $d^{\frac{1}{n}} r -r(1+r)/2$ and the remain $n-1$ MPS takes $(n-1)(d^{\frac{1}{n}} r^2 -r(1+r)/2)$.%And the compression ratio is 

We also consider a metric of normalized storage, {\em compression ratio}, which is the  ratio of subspace storage to the entire $Nd$ amount of data storage, or equivalently 
%\begin{equation}
$\rho_{\text{ST}} = \frac{\text{dim(ST)}}{N_\text{tr}d},$ 
%\end{equation}
where ST can be any of PCA, T-PCA, or TT-PCA. 

\noindent {\bf Computation Complexity of finding reduced subspace:} We will now find the complexity of the three PCA algorithms (standard PCA, T-PCA, and TT-PCA). 
We assume that there are $C$ classes,  $N_\text{tr}$ is the total number of training data points, and $N_\text{te}$ be the total number of test data points.
To compute standard PCA, we first compute the covariance matrix of the data, whose complexity is $O(d^2 N_\text{tr})$. This is followed by eigenvalue decomposition of the covariance matrix, whose complexity is $O(d^3)$. Thus, the overall complexity is $O(d^2\max(N_\text{tr},d))$.  
To compute the subspace corresponding to T-PCA, we  first  compute  $n$ orthonormal linear transformations using SVD, which takes $O(nd^{\frac{1}{n}} r^2)$ \cite{oseledets2011tensor} time. This is followed by finding the  subspace basis for the dimensional reduced tensor by PCA, which takes $O( r^{2n}\max(N_\text{tr}, r^n))$ time. Thus, the total computation complexity is $O(r^{2n}\max(N_\text{tr}, r^n) + nd^{\frac{1}{n}} r^2)$. 
The computation complexity for finding the tensor train subspace needs the recovery of the $n$ components ($\mathscr{U}_1, \cdots,  \mathscr{U}_n$), which takes $O(nd^{\frac{1}{n}}r^3)$ time  for calculation based on Algorithm \ref{TTPCA_Algo}.

\noindent {\bf Classification Complexity: }  Prediction under standard PCA  model is equivalent to solving \eqref{eq: pred}, whose computation complexity is $O(N_\text{te} C dr)$.  
For T-PCA, we need additional step to make ${\bf U}$ for each class, which required an additional complexity of $O(dC r^2)$. %Note that the computational time is also tensor order dependent where the effect is stored in $d$. 
Thus, the overall complexity for prediction based on T-PCA is $O( C dr\max(N_\text{te}, r))$. 
TT-PCA needs the same steps as T-PCA where first a conversion to ${\bf U}$ is needed which has a complexity of $O(dC r^2)$ for each class giving an overall complexity of  $O( C dr\max({N}_\text{te}, r))$. 

%Making prediction needs to recover the subspace from $n$ linear transformation and the $r$ core tensors, which takes an extra $O(dr^2)$ as compared with PCA. Thus the total complexity in making prediction is $O(dr^2 + \tilde{N}Cdr)$, where the term $O(\tilde{N}Cdr)$ dominates the complexity.

%The prediction process for TT-PCA requires the process for recovering the subspace from the MPS, which is similar to T-PCA that takes $O(dr^2)$ for computing. Thus, the total computation complexity for prediction is  $O(dr^2 + \tilde{N}Cdr)$.

These results for  PCA, T-PCA and TT-PCA are summarized in Table \ref{ComplexityTable}, where the lowest complexity entries in each column are bold-faced. 
We can see that TT-PCA has advantages in both storage and subspace computation. Although TT-PCA degrades in computation complexity compared with PCA in making prediction, the extra complexity is dependent on amount of testing data and is negligible for $N_\text{te}>r$.

\begin{table*}[t!]
  \centering
  \scalebox{1}{
  \begin{tabular}{cccc}
    \toprule
    & Storage & Subspace Computation & Classification\\
    \midrule
    PCA 	&  $dr-\frac{r(1+r)}{2}$	& $O(d^2\max(N_\text{tr}, d))$ 		& ${\bf O(CdrN_\text{te})}$\\
    T-PCA	&  $r^{n+1} + n(d^{\frac{1}{n}}r-\frac{r(r+1)}{2})$ 	& $O(r^{2n}\max(N_\text{tr} + r^n) + nd^{\frac{1}{n}} r^2)$ 	& $O( C dr\max({N}_\text{te}, r))$\\
    TT-PCA&  ${\bf d^{\frac{1}{n}} r(r(n-1) +1) -\frac{r(r+1)n}{2}}$ & ${\bf O(nd^{\frac{1}{n}} r^3)}$ 				& $O( C dr\max({N}_\text{te}, r))$\\
    \bottomrule
  \end{tabular}
  }
   \label{ComplexityTable}
   \vspace{0.1in}
  \caption{\small Storage and computation complexity for PCA algorithm. The bold entry in each column depicts the lowest order.  }
   \vspace{-.2in}
\end{table*}

\section{Experimental Results}\label{simu}
%\vspace{-.05in}
%\subsection{TT-PCA on YaleFace Dataset}
In this section, we compare the proposed TT-PCA algorithm with the T-PCA  \cite{lu2011survey,yan2007multilinear, lu2009uncorrelated}, and the standard PCA algorithms.
%PCA is selected as it is the predominant subspace model in data analysis. 
T-PCA is a Tucker decomposition based PCA that has been shown to be effective in face recognition. The evaluation is conducted in the Extended YaleFace Dateset B \cite{georghiades2001few, lee2005acquiring}, which consists of  38 persons with 64 faces each that are taken under different illumination, where each face is represented by a matrix of size $48 \times 42$. 
%Thus,  the dataset formed by the images of 38 people with 64 faces each can be represented by  a tensor of size ${48 \times 42 \times 64 \times 38}$. 
Each element of the face is the grayscale intensity of the pixel which can have value from 0 to 255.  Extended YaleFace Dateset B has been shown to satisfy subspace structure \cite{basri2003lambertian}, which motivates our choice for exploring multi-dimensional subspace structures in this dataset. 
%We further reshape each face into a $4^{\text{th}}$ order tensor of size $\mathbb{R}^{6 \times 8 \times 6 \times 7}$ for tensor analysis.
%We further reshape the matrix of size  $48 \times 42$ into a $4^{\text{th}}$ order tensor of size ${6\times 8 \times 6 \times 7 }$ to test the approach for larger orders. 
\emph{For the experiments each image of a person is reshaped as $\mathscr{X}_{i}\in \mathbb{R}^{6\times 8 \times 6 \times 7 }$ to validate the approach using tensor subspaces.}%,  and a Gaussian noise  $\mathcal{N}(0, 100)$ is added to each pixel of the data. 

% for high order tensor analysis. 

%We let  $\mathscr{X}_{i}\in \mathbb{R}^{6\times 8 \times 6 \times 7 }$ be the image of a face from one person and $N_\text{tr}=20$ images are selected for the person. 
%In other words, we choose 20 images of one person to make a sub-dataset for performing PCA. 
%Before performing PCA, we add a Gaussian noise $\mathcal{N}(0, 900)$ to each pixel of the  image. 
%We first consider the dominant eigen-face for the three approaches. 
%For TT-PCA and T-PCA approach, we have the flexibility to select the value of threshold $\tau$. 
%In Figure \ref{PCA_YaleFaceEigen}, we compare the eigen-faces for PCA, T-PCA, and TT-PCA for different values of $\tau$. 
%From left to right in the T-PCA row, we select $\tau$ as $0, 0.05, 0.07, 0.09, 0.11, 0.13, 0.15, 0.17$ while for TT-PCA, we select $\tau$ as $0, 0.008, 0.025, 0.04, 0.06, 0.08, 0.1, 0.22$. The compression ratios are shown above the image in Figure \ref{PCA_YaleFaceEigen}. 
%We note that TT-PCA is able to get more features of the person in the dominant image. 
%As $\tau$ increases, there is lower tensor rank and this more structure in the blocks are used (due to image reshaping) which leads to performance degradation. 
%However, at similar compression ratios, TT-PCA seems better than T-PCA. 
%there is lower tensor rank and this more structure in the blocks are used (due to image reshaping) which leads to performance degradation. 

\begin{figure}[t!] 
\includegraphics [trim=0.5in 0in 1in 0in, keepaspectratio, width=0.4\textwidth] {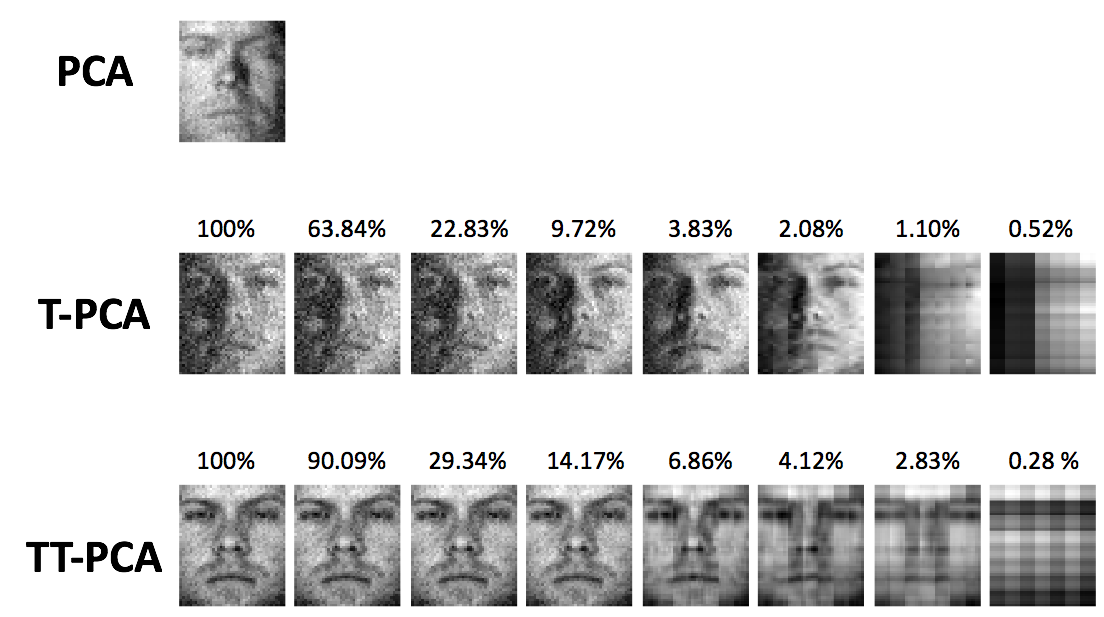}
\centering
\caption{\small First Eigen Face for PCA and First Tensor Face for T-PCA and TT-PCA under different compression ratios. The number at top are the compression ratios. }
\label{PCA_YaleFaceEigen}
\vspace{-.15in}
\end{figure}

\textbf{Data representation via tensor subspaces}- We first compare the first dominant eigen-face for PCA  and the first dominant tensor-face for T-PCA and TT-PCA by sampling $N_\text{tr} =20$ images from one randomly selected person, and reshape each of the images into a $4^\text{th}$ order tensor for tensor PCA analysis. 
We add a Gaussian noise $\mathcal{N}(0, 900)$ to each pixel of the image. TT-PCA and T-PCA have the flexibility in controlling compression ratio by switching $\tau$, where larger $\tau$ gives high compression ratio and less accuracy in approximation and vice versa. 
Figure \ref{PCA_YaleFaceEigen} shows the tensor-face for T-PCA and TT-PCA under different compression ratio, and the eigen-face for PCA, where the compression ratio (marked on top) is decreasing from left-right (implying increasing data compression) for the tensor PCA algorithms.  TT-PCA shows a better performance in constructing tensor-face than both the T-PCA and PCA algorithms  since the dominant eigen-face pictorially takes more features of the noiseless image of the person. %looks most like the original face. %tensor face constructs a tensor-face that is less subjective to noise and illumination. 
As $\tau$ increases (which changes compression, from left to right), tensor rank becomes lower and tensor-face degrades to more blurry images. %However, the small blocks in tensor-faces under lower compression ratio indicates that tensor train structure effectively captures the structure in the faces. 
Under similar compression ratio, such as $6.86\%$ for TT-PCA and $9.72\%$ for T-PCA, TT-PCA performs better than T-PCA since the tensor face is less affected by  noise.

\begin{figure}[t!] 
	\includegraphics [trim=0.5in 0in 1in 0in, keepaspectratio, width=0.4\textwidth] {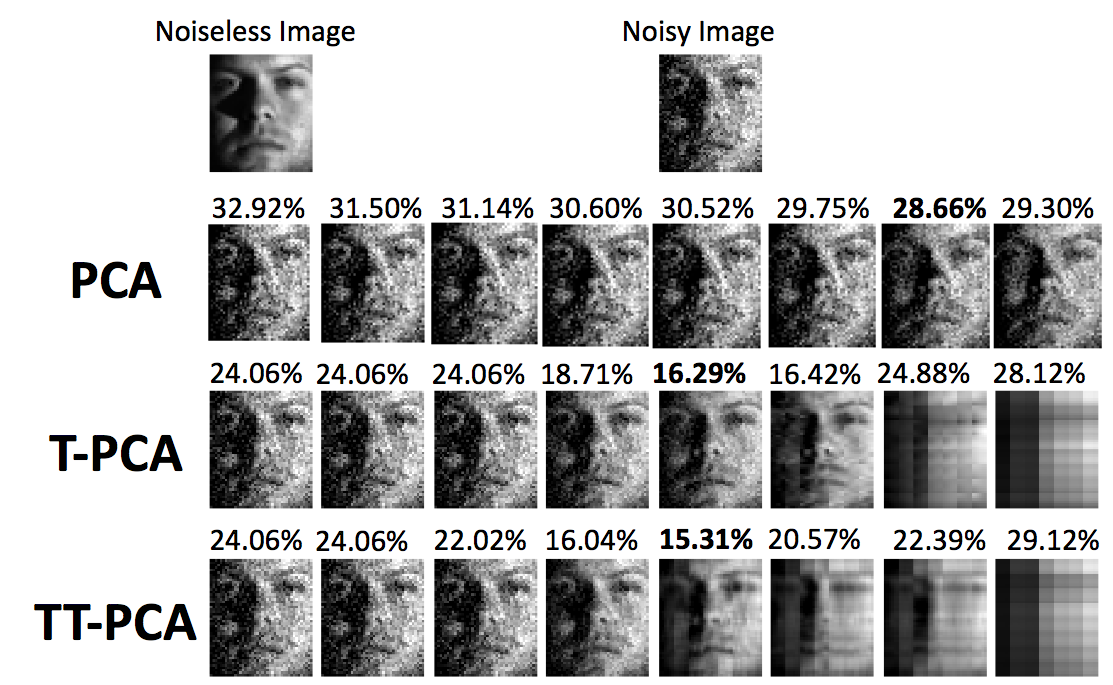}
	\centering
	\caption{\small Face denoising under PCA, T-PCA and TT-PCA. The reconstruction errors  are marked on top of each image. Different images in each row correspond to decreasing compression ratios (increasing compression, increasing $\tau$) from left to right. The compression ratios for T-PCA and TT-PCA are the same (left-right) as that in Fig. \ref{PCA_YaleFaceEigen}.}
	\label{Denoise}
	\vspace{-.15in}
\end{figure}

We further illustrate one image sampled from the $20$ noisy images and its projection onto 
(a) the linear subspace given by PCA with ranks being $ 16, 14, 12, 10, 8, 6, 4, 2$ (from left-to-right), which gives compression ratios of $0.8, 0.7, 0.6, 0.5, 0.4, 0.3, 0.2, 0.1$, 
(b) the multi-linear subspace given by T-PCA with compression ratio $1, 0.638. 0.228, 0.972, 0.038, 0.021, 0.011, 0.005$, 
and (c) tensor train subspace given by TT-PCA with compression ratio $1, 0.901, 0.293, 0.142, 0.069, 0.041,0.028, 0.003$. 
The reconstruction error, defined as the distance between the original image (without noise) and the projection of the noisy image to the subspace, is depicted at the top of images in Figure \ref{Denoise}. As seen from the figure the reconstruction errors of T-PCA and TT-PCA are significantly lower than that of PCA, and TT-PCA gives the lowest $15.31\%$ reconstruction error under $0.069$ compression ratio.

\begin{figure}[t!] 
	\includegraphics [trim=0.5in 2.3in 1in 2.8in, keepaspectratio, width=0.22\textwidth] {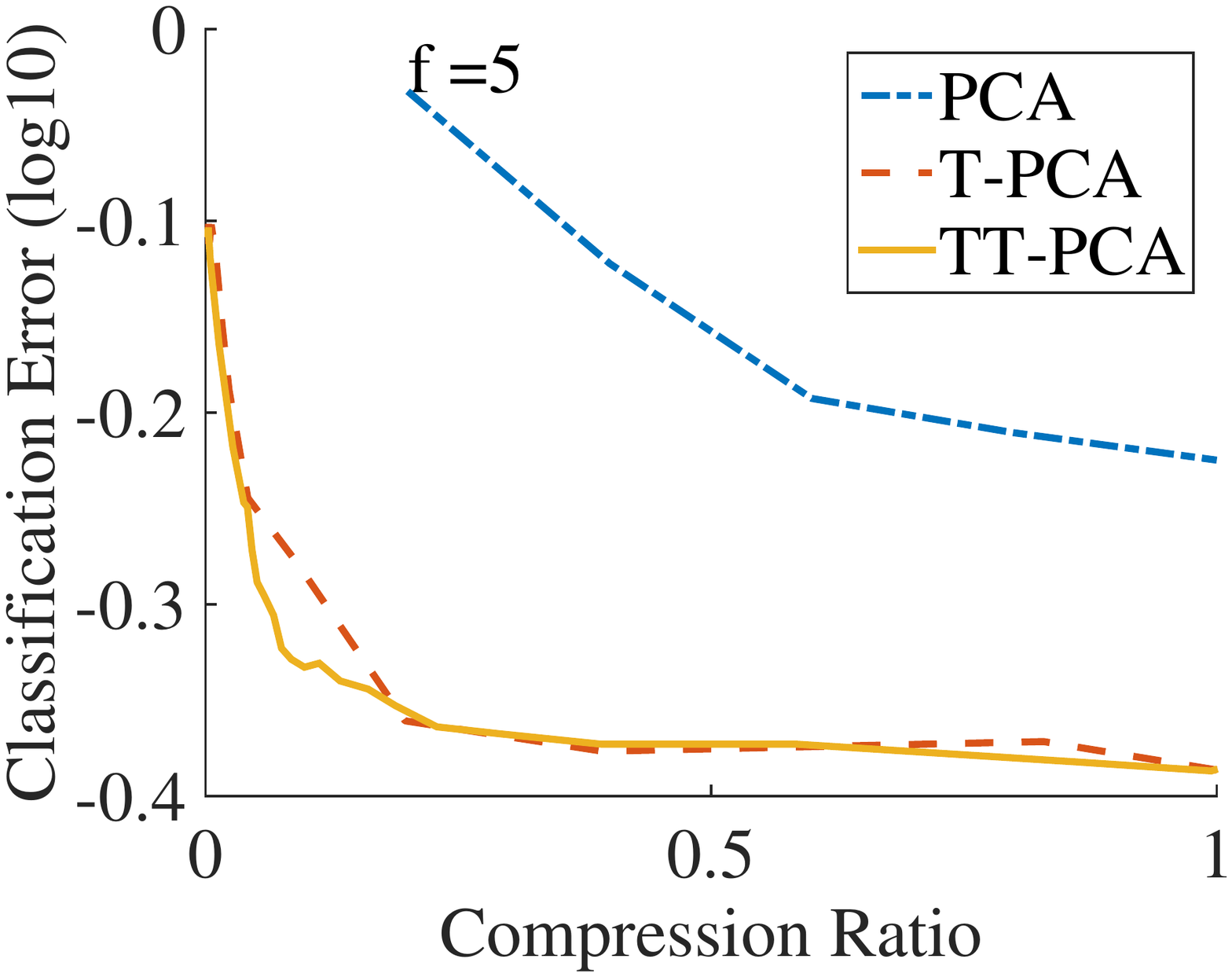}
	\includegraphics [trim=0.5in 2.3in 1in 2.8in, keepaspectratio, width=0.22\textwidth] {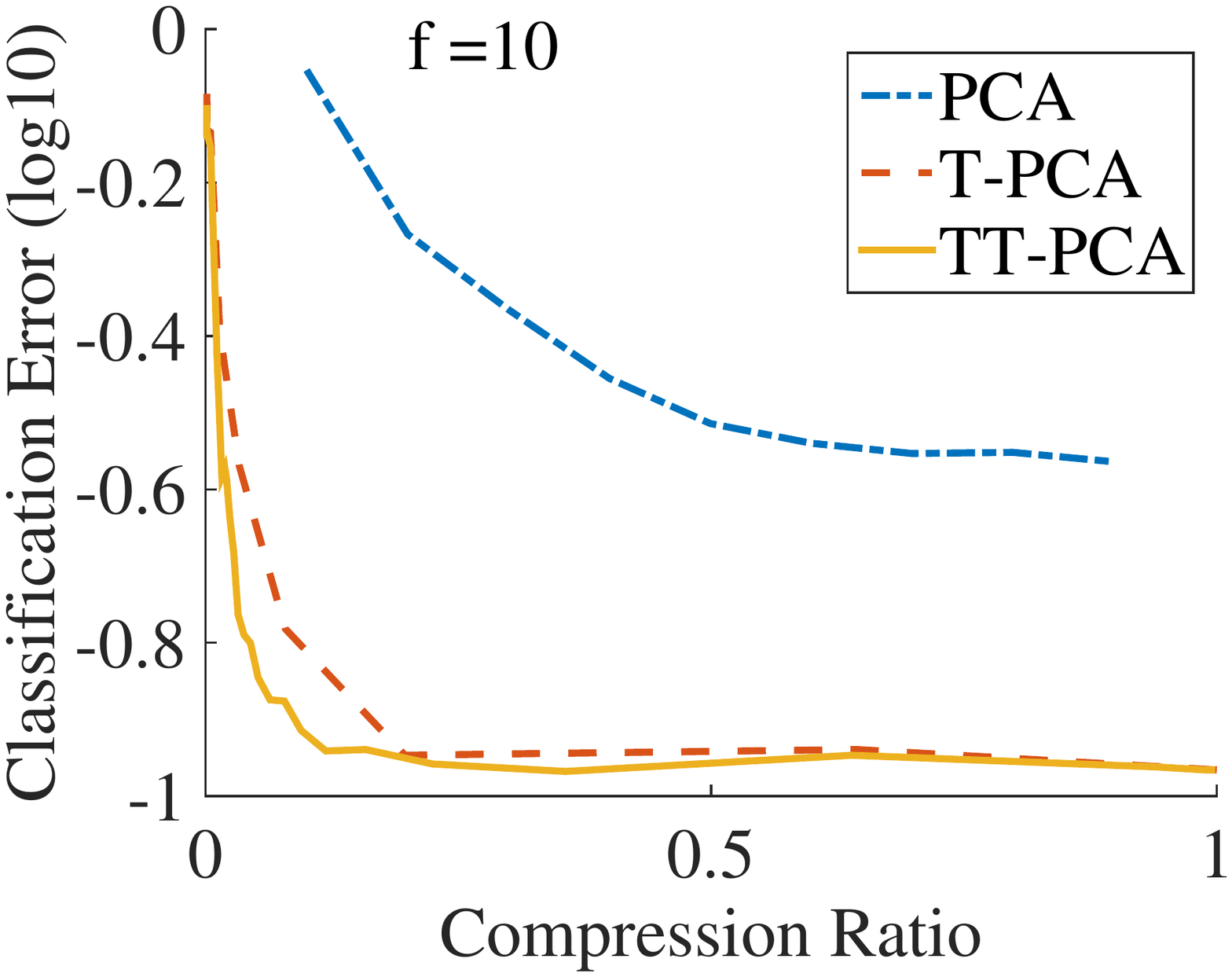}
	\includegraphics [trim=0.5in 2.3in 1in 2.8in, keepaspectratio, width=0.22\textwidth] {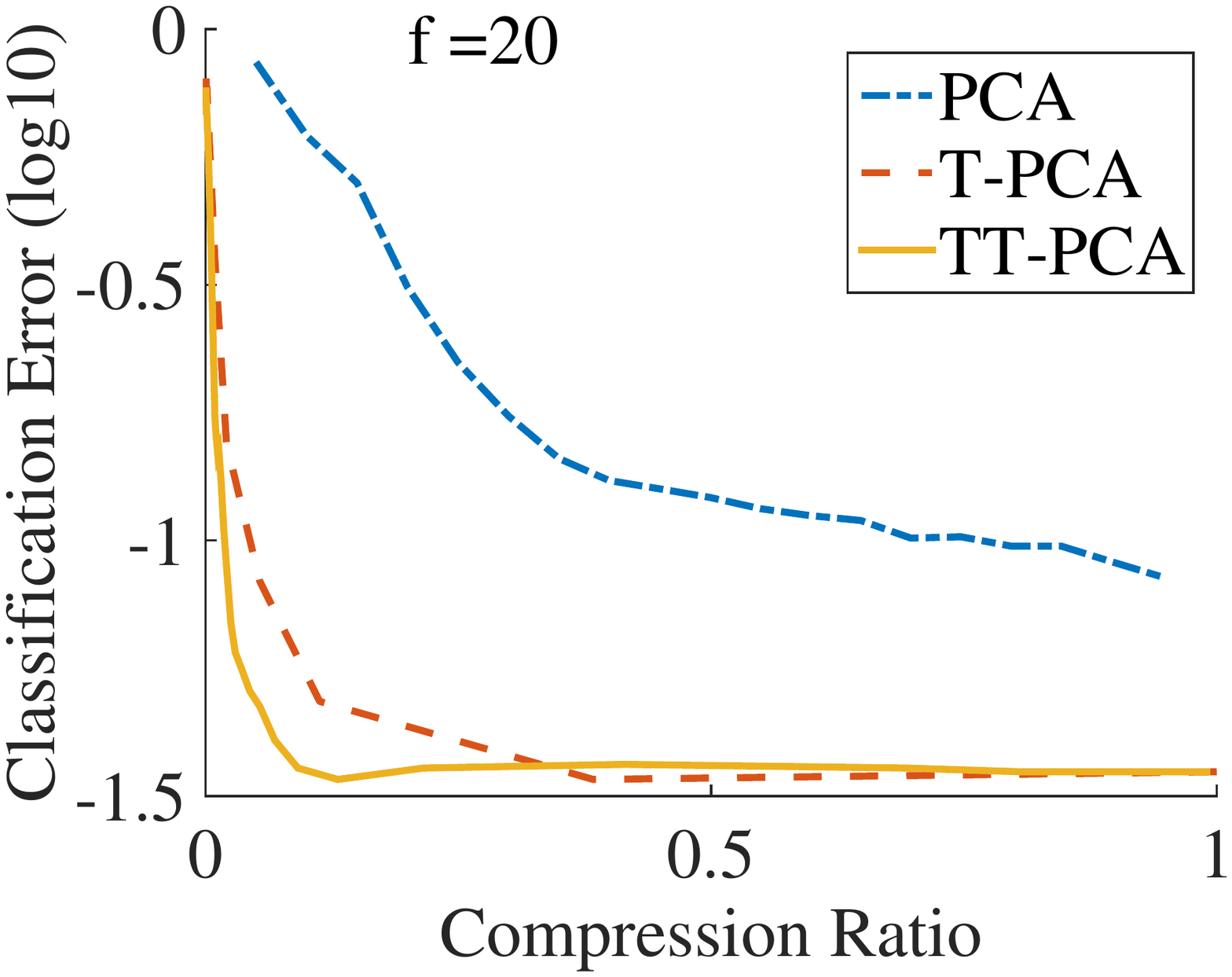}
	\includegraphics [trim=0.5in 2.3in 1in 2.8in, keepaspectratio, width=0.22\textwidth] {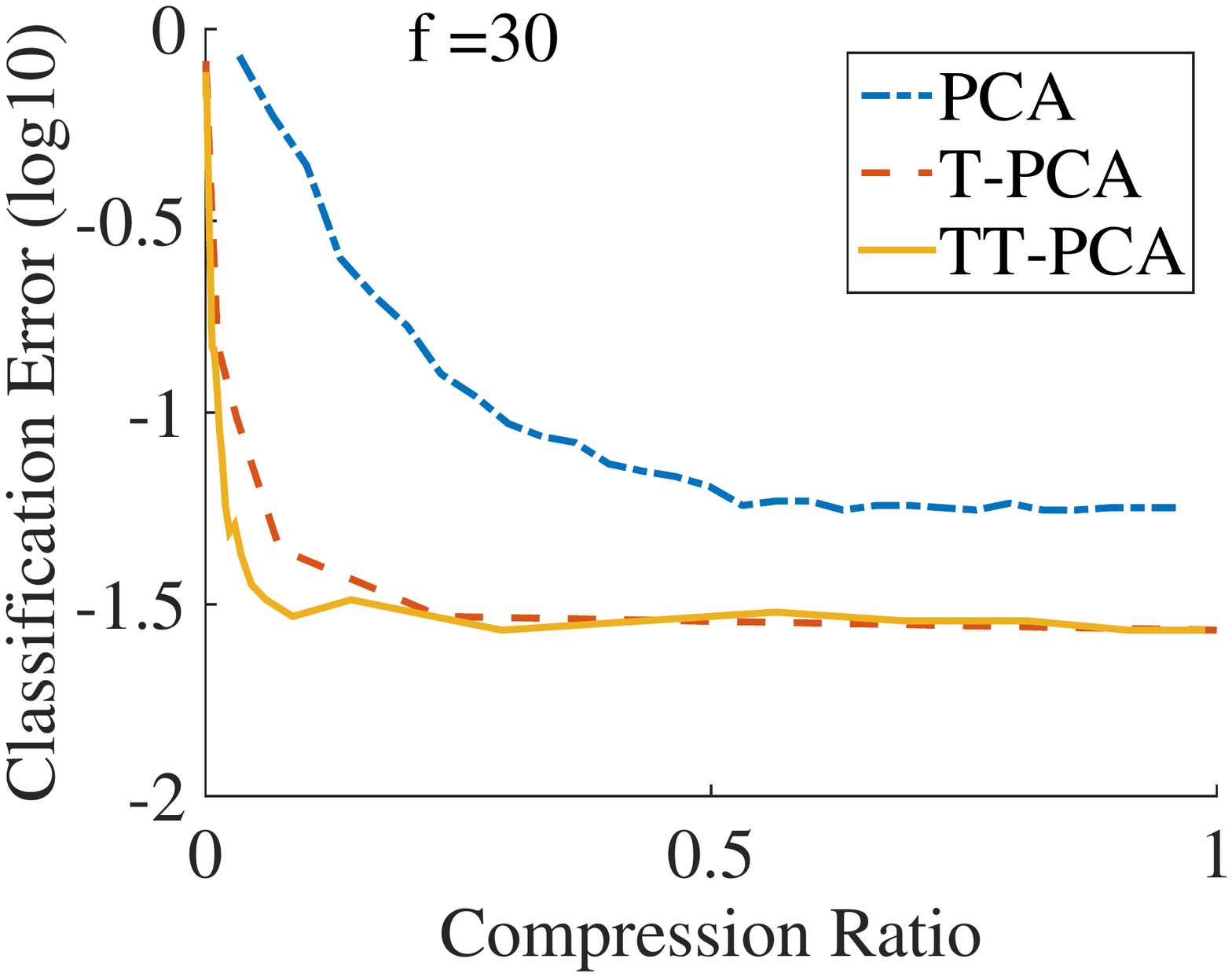}
	\centering
	%\vspace{-.05in}
	\caption{\small Clustering Error in $\log10$ scale versus Compressed Ratio for Extended YaleFace Dataset B Dataset. 38 faces with noise are selected from the data set and the training sample size is 5, 10, 20, 30 (from left to right, top to bottom) respectively. %The best performance results of PCA, T-PCA and TT-PCA are shown in the Table.
		}
	\label{PCA_YaleFace}
	\vspace{-.15in}
\end{figure}

\begin{table}[t!] 
	\includegraphics [trim=0.5in 0in 1in -0.2in, keepaspectratio, width=0.4\textwidth] {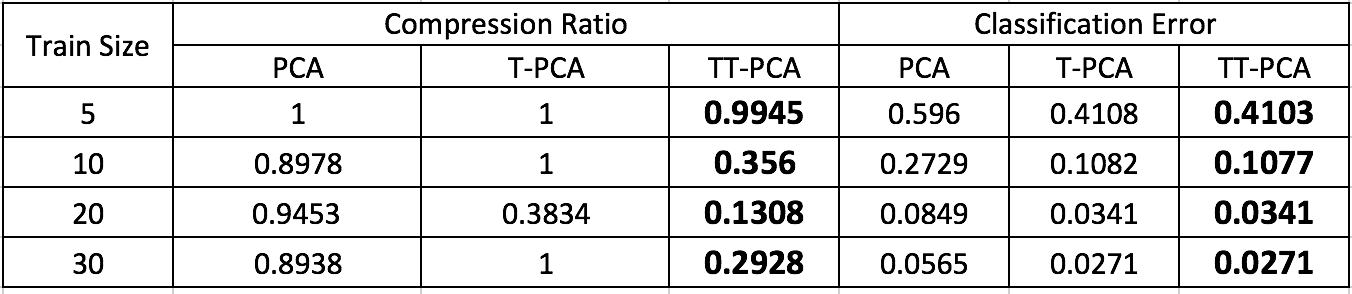}
	\centering
	\caption{\small Compression ratio and classification error at the lowest classification error}
	\label{PCA_YaleFaceTable}
	\vspace{-.15in}
\end{table}

\textbf{Classification using TT-PCA} - Next, we test the performance of the three approaches for classification. For classification, we choose $f$ training data points (at random) from each of the $38$ people, and thus the amount of training data points is $N_\text{tr}=38f$. 
The remaining data of each person is used for testing, and thus $N_\text{te}= 38(64-f)$. We add a Gaussian noise  $\mathcal{N}(0, 100)$ to each pixel of the data.
%We let $f=5, 10, 20, 30$. 
%The selection of rank gives the flexibility for compressing data for standard PCA, while the thresholding parameter $\tau$ in T-PCA and TT-PCA gives the flexibility for compressing data. 
For training sizes $f=5, 10, 20, 30$, Figure \ref{PCA_YaleFace} compares the classification error of the different algorithms as a function of compression ratio for each $f$. We note that as $f$ increases, the classification performance becomes better for all algorithms. We further see that TT-PCA performs better at low compression ratios, and the classification error increases after first decreasing. This is because with higher compression ratios (low compression), the approaches will try to over-fit noise leading to lower classification accuracy. 

Table \ref{PCA_YaleFaceTable} highlights the data from Figure \ref{PCA_YaleFace} to illustrate the improved performance of TT-PCA. 
This table shows the compression ratio at which the best classification performance is achieved, and the classification error at this compression ratio. We note that the point at which best compression ratio is achieved is lowest for TT-PCA, and so is the best classification error thus demonstrating that TT-PCA is able to extract the data structure well at high data compressions. \emph{This indicates that human face data under different illumination conditions lies not only close to the subspace models, but are better approximated by tensor train subspace models. Further we note that TT-PCA requires far less training sample size compared to other approaches.}

\section{Conclusion}\label{concl}

This paper outlines novel algorithms and methods for tensor train subspaces for data representation. A PCA like algorithm namely TT-PCA is proposed. This algorithm is validated on vision dataset and exhibit improved classification performance, better dimensionality reduction, and lower computational complexity as compared to the considered baseline approaches.

{\small
\bibliographystyle{ieee}
\bibliography{Ref}
}

\end{document}